\documentclass[articlekbc,twoside,11pt]{article}
\usepackage{akbc}
\akbcheading
\newcommand\mytitle{OPIEC: An Open Information Extraction Corpus}
\ShortHeadings{\mytitle}
{Gashteovski, Wanner, Hertling, Broscheit \& Gemulla}
\finalcopy 

\usepackage{hyperref}
\usepackage[usenames, dvipsnames]{color}
\usepackage{algorithm}
\usepackage[noend]{algpseudocode}
\usepackage{mathtools}
\usepackage{tikz}
\usepackage{tikz-dependency}
\usetikzlibrary{backgrounds}
\usepackage{pgf}
\usepackage{adjustbox}
\usepackage{longtable}
\usepackage{array}
\usepackage{footnote}
\makesavenoteenv{tabular}
\makesavenoteenv{table}
\usepackage{enumitem}

\newcommand{\ignore}[1]{}

\definecolor{forestgreen}{rgb}{0.13, 0.55, 0.13}

\usepackage{pifont}
\newcommand{\cmark}{\ding{51}}%
\newcommand{\xmark}{\hspace{.27em}-\hspace{.27em}}%

\definecolor{mydarkgreen}{rgb}{0.0,0.5,0.0}

\usepackage{booktabs}
\usepackage{subcaption}
\begin{document}

\title{\mytitle}

\author{\name Kiril Gashteovski \email k.gashteovski@uni-mannheim.de \\
	\name Sebastian Wanner \email swanner@mail.uni-mannheim.de \\
        \name Sven Hertling \email sven@informatik.uni-mannheim.de \\
        \name Samuel Broscheit \email broscheit@informatik.uni-mannheim.de \\
        \name Rainer Gemulla \email rgemulla@uni-mannheim.de \\
       \addr Universit{\"a}t Mannheim, Germany
       }

\maketitle

\begin{abstract}
  Open information extraction (OIE) systems extract relations and their
  arguments from natural language text in an unsupervised manner. The resulting
  extractions are a valuable resource for downstream tasks such as knowledge
  base construction, open question answering, or event schema induction. In this
  paper, we release, describe, and analyze an OIE corpus called OPIEC, which was
  extracted from the text of English Wikipedia. OPIEC complements the available
  OIE resources: It is the largest OIE corpus publicly available to date (over
  340M triples) and contains valuable metadata such as provenance information,
  confidence scores, linguistic annotations, and semantic annotations including
  spatial and temporal information. We analyze the OPIEC corpus by comparing its
  content with knowledge bases such as DBpedia or YAGO, which are also based on
  Wikipedia. We found that most of the facts between entities present in OPIEC
  cannot be found in DBpedia and/or YAGO, that OIE
  facts 
  often differ in the level of specificity compared to knowledge base facts, and
  that OIE open relations are generally highly polysemous. We believe that the
  OPIEC corpus is a valuable resource for future research on automated knowledge
  base construction.
\end{abstract}


\section{Introduction}

Open information extraction (OIE) is the task of extracting relations and their
arguments from natural language text in an unsupervised manner
\cite{banko2007open}. The output of such systems is usually structured in the
form of (\emph{subject}, \emph{relation}, \emph{object})-triples. For example,
from the sentence \textit{``Bell is a telecommunication company, which is based
  in L. A.,''} an OIE system may yield the extractions \textit{(``Bell'';
  ``is''; ``telecommunication company'')} and \textit{(``Bell''; ``is based
  in''; ``L. A.'')}. The extractions of OIE systems from large corpora are a
valuable resource for downstream tasks~\cite{etzioni2008open,mausam2016open}
such as automated knowledge base
construction~\cite{riedel2013relation,wu2018towards,vashishth2018cesi,shi2018open},
open question answering~\cite{fader2013paraphrase}, event schema
induction~\cite{balasubramanian2013generating}, generating inference
rules~\cite{jain2016knowledge}, or for improving OIE systems
themselves~\cite{schmitz2012open,yahya2014renoun}. A number of derived resources
have been produced from OIE extractions, including as entailment
rules~\cite{jain2016knowledge}, question paraphrases~\cite{fader2013paraphrase},
Rel-grams~\cite{balasubramanian2012rel}, and OIE-based
embeddings~\cite{stanovsky2015open}.

In this paper, we release a new OIE corpus called \emph{OPIEC}.\footnote{The
  OPIEC corpus is available at
  \url{https://www.uni-mannheim.de/dws/research/resources/opiec/}.} The OPIEC
corpus has been extracted from the full text of the English Wikipedia using the
Stanford CoreNLP pipeline~\cite{manning2014stanford} and the state-of-the-art
OIE system MinIE~\cite{gashteovski2017minie}. OPIEC complements available OIE
resources~\cite{fader2011identifying,lin2012entity,nakashole2012patty,moro2012wisenet,moro2013integrating,bovi2015large,delli2015knowledge}:
It is the largest OIE corpus publicly available to date (with over 340M triples)
and contains valuable metadata information for each of its extractions not
available in existing resources (see Tab.~\ref{tab:oie_corpora} for an
overview). In particular, OPIEC provides for each triple detailed provenance
information, syntactic annotations (such as POS tags, lemmas, dependency
parses), semantic annotations (such as polarity, modality, attribution, space,
time), entity annotations (NER types and, when available, Wikipedia links), as
well as confidence scores.

We performed a detailed data profiling study of the OPIEC corpus to analyze its
contents and potential usefulness for downstream applications. We observed that
a substantial fraction of the OIE extractions was not self-contained (e.g.,
because no anaphora resolution was performed) or overly specific (e.g., because
arguments were complex phrases). Since these extractions are more difficult to
work with, we created the \textit{OPIEC-Clean} subcorpus (104M triples), in
which we only retained triples that express relations between concepts. In
particular, OPIEC-Clean contains triples in which arguments are either named
entities (as recognized by an NER system), match a Wikipedia page title (e.g.,
concepts such as \emph{political party} or \emph{movie}), or link directly to a
Wikipedia page. Although OPIEC-Clean is substantially smaller than the full
OPIEC corpus, it is nevertheless four times larger than the largest prior OIE
corpus.

To gain insight into the information present in the OPIEC corpus, we compared
its content with the DBpedia~\cite{bizer2009dbpedia} and YAGO~\cite{hoffart2013yago2} knowledge bases,
which are also constructed from Wikipedia (e.g., from infoboxes). Since such an
analysis is difficult to perform due to the openness and ambiguity of OIE
extractions, we followed standard practice and used a simple form of distant
supervision. In particular, we analyze the \emph{OPIEC-Linked} subcorpus (5.8M
triples), which contains only those triples in which both arguments are linked
to Wikipedia articles, i.e., where we have golden labels for disambiguation. We
found that most of the facts between entities present in OPIEC-Linked cannot be
found in DBpedia and/or YAGO, that OIE facts often differ in the level of specificity compared to 
knowledge base facts, and that frequent OIE open relations are generally
highly polysemous.

Along with the OPIEC corpus as well as the OPIEC-Clean and OPIEC-Linked
subcorpora, we release the codebase used to construct the corpus as well as a
number of derived resources, most notably a corpus of open relations between 
arguments of various entity types along
with their frequencies. We believe that the OPIEC corpus is a valuable resource
for future research on automated knowledge base construction.


\section{Related OIE Corpora}

Many structured information resources created from semi-structured or
unstructured data have been constructed in recent years. Here we focus on
large-scale OIE corpora, which do not make use of a predefined set of arguments
and/or relations. OIE corpora complement more targeted resources such as
knowledge bases (e.g., DBpedia, YAGO) or NELL~\cite{mitchell2018never}, as well
as smaller, manually crafted corpora such as the one of
\citet{stanovsky2016creating}; see also the discussion in
Sec.~\ref{sec:analysis}. An overview of the OPIEC corpus, its subcorpora, and
related OIE corpora is given in Tab.~\ref{tab:oie_corpora}.

\begin{table}
	\scriptsize
	\centering
	\begin{tabular}{lrrrccccc} \toprule
                    & \# triples & \# unique & \# unique & disamb. args& confi-	& prove- & syntactic   & semantic    \\ 
                    &  (millions)          & arguments & relations & (aut./gold) & dence	& nance	 & annotat. & annotat. \\
            &        & (millions)           & (millions)& &  & & &   \\ \midrule
		ReVerb          & 14.7       & 2.2       & 0.7       & \xmark{} / \xmark         & \cmark	& \cmark & \cmark      & \xmark      \\ 
		ReVerb-Linked   & 3.0        & 0.8       & 0.5       & \cmark{} / \xmark         & \xmark	& \xmark & \xmark      & \xmark      \\ 
		PATTY (Wiki)    & 15.8       & 0.9       & 1.6       & \cmark{} / \xmark         & \xmark	& \xmark & \xmark      & \xmark      \\ 
		WiseNet 2.0     & 2.3        & 1.4       & 0.2       & \xmark{} / \cmark         & \xmark	& \xmark & \xmark      & \xmark      \\ 
		DefIE           & 20.3       & 2.5       & 0.3       & \cmark{} / \xmark         & \xmark	& \xmark & \xmark      & \xmark      \\ 
		KB-Unify        & 25.5       & 2.1       & 2.3       & \cmark{} / \xmark         & \xmark	& \xmark & \xmark      & \xmark      \\ \midrule
		OPIEC           & 341.0      & 104.9     & 63.9      & \xmark{} / \xmark         & \cmark & \cmark & \cmark      & \cmark      \\ 
		OPIEC-Clean     & 104.0      & 11.1      & 22.8      & \xmark{} / \xmark         & \cmark	& \cmark & \cmark      & \cmark      \\
		OPIEC-Linked    & 5.8        & 2.1       & 0.9       & \xmark{} / \cmark         & \cmark	& \cmark & \cmark      & \cmark      \\ \bottomrule
 	\end{tabular}
	\caption{Available OIE corpora and their properties. All numbers are in
    millions. Syntactic annotations include POS tags, lemmas, and dependency
    parses. Semantic annotations include attribution, polarity, modality, space,
    and time.}
	\label{tab:oie_corpora}
\end{table}

One of the first and widely-used OIE resource is the ReVerb
corpus~\cite{fader2011identifying}, which consists of the high-confidence
extractions of the ReVerb OIE system from the ClueWeb09 corpus. In subsequent
work, a subset of the ReVerb corpus with automatically disambiguated arguments
was released~\cite{lin2012entity}. The PATTY~\cite{nakashole2012patty},
WiseNet~\cite{moro2012wisenet}, WiseNet 2.0~\cite{moro2013integrating}, and
DefIE~\cite{bovi2015large} corpora additionally organize open relations in
relational synsets and then the relational synsets into taxonomies. Finally,
KB-Unify~\cite{delli2015knowledge} integrates multiple different OIE corpora
into a single resource. 

The largest prior corpus is KB-Unify, which consists of 25.5M triples in roughly
4.5M distinct open relations and arguments. Both OPIEC (341M and 105M/64M, resp.)
and OPIEC-Clean (104M and 11M/23M) are significantly larger than KB-Unify, both in
terms of number of triples as well as in terms of distinct arguments and
relations. One of the reasons for this size difference is that the MinIE
extractor, which we used to create the OPIEC corpus, produces more extractions
than the extractors used to create prior resources. The OPIEC corpus---but not
OPIEC-Clean and OPIEC-Linked--also contains all extractions produced by MinIE
unfiltered, whereas most prior corpora use filtering rules aiming to provide
higher-quality extractions (e.g., frequency constraints).

Most of the available corpora use automated methods to disambiguate entities
(e.g., w.r.t.~to a knowledge base). On the one hand, such links are very useful
because ambiguity is restricted to open relations. On the other hand, the use
of automated entity linkers may introduce errors and---perhaps more
importantly---restricts the corpus to arguments that can be confidently linked. We
did not perform automatic disambiguation in OPIEC, although we retained
Wikipedia links when present (similar to WiseNet). Since these links are
provided by humans, we consider them as golden disambiguation links. The
OPIEC-Linked subcorpus contains almost 6M triples from OPIEC in which both
arguments are disambiguated via such golden links.

A key difference between OPIEC and prior resources is in the amount of metadata
provided for each triple. First, only ReVerb and OPIEC provide confidence scores
for the extractions. The confidence score measures how likely it is that the
triple has been extracted correctly (but not whether it is actually true). For
example, given the sentence \textit{``Bill Gates is a founder of the
  Microsoft.''}, the extraction \textit{(``Bill Gates''; ``is founder of'';
  ``Microsoft'')} is correct, whereas the extraction \textit{(``Bill Gates'';
  ``is''; ``Microsoft'')} is not. Since OIE extractors are bound to make
extraction errors, the extractor confidence is an important signal for
downstream applications~\cite{dong2014knowledge}. Similarly, provenance
information (i.e., information about where the triple was extracted) is not
provided in many prior resources.

One of the reasons why we chose MinIE to construct OPIEC is that it provides
syntactic and semantics annotations for its extractions. Syntactic annotations
include part-of-speech tags, lemmas, and dependency parses. Semantic annotations
include attribution (source of information according to sentence), polarity
(positive or negative), and modality (certainty or possibility). We also extended
MinIE to additionally provide spatial and temporal annotations for its triples.
The use of semantic annotations simplifies the resulting triples significantly
and provides valuable contextual information; see the discussion in
Section~\ref{sec:construction}.


\section{Corpus Construction}
\label{sec:construction}

\begin{figure}
    \centering
    \includegraphics[width=0.95\textwidth]{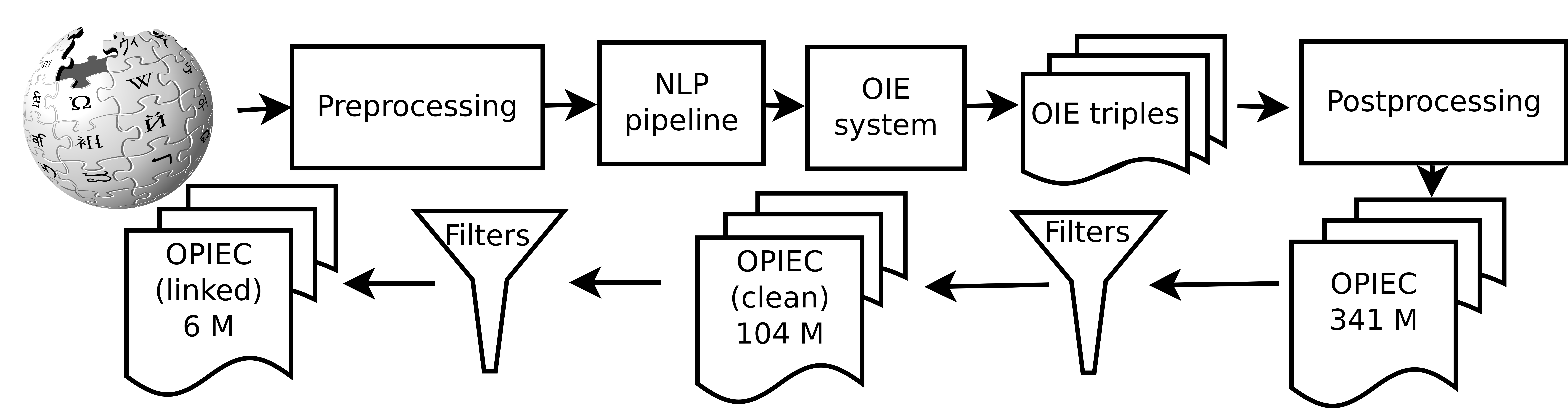}
    \caption{Corpus construction pipeline} 
    \label{fig:pipeline}
\end{figure}

OPIEC was constructed from all the articles of the English Wikipedia dump of
June 21, 2017. Fig.~\ref{fig:pipeline} gives an overview of the pipeline that we
used. The pipeline is (apart from preprocessing) not specific to Wikipedia and
thus can be used with other dataset as well. We used Apache Spark \cite{zaharia2016apache} to distribute
corpus construction across a compute cluster so that large datasets can be
handled. We release the entire codebase along with the actual OPIEC corpora.

\subsection{Preprocessing} 

We used a modified version of
WikiExtractor\footnote{\url{https://github.com/attardi/wikiextractor}} to
extract the plain text from Wikipedia pages. In particular, we modified
WikiExtractor such that it retains internal Wikipedia links to other Wikipedia
articles. The links are provided as additional metadata in the form of
(\textit{span}, \textit{target page}) annotations. Custom entity linkers can be
inserted into the pipeline by providing such annotations.

Wikipedia generally does not link the first phrase of an article to the article
page itself; e.g., the page on New Hampshire starts with \textit{``New Hampshire
  is a ...''}, where \textit{New Hampshire} is not linked. To avoid losing this
relationship, we link the first phrase that \emph{exactly} matches the Wikipedia
page name (if any) to that Wikipedia page. 



\subsection{NLP Pipeline} 

We ran an NLP pipeline on the preprocessed Wikipedia articles by using
CoreNLP~\cite{manning2014stanford}, version 3.8.0. We performed tokenization,
sentence splitting, part-of-speech tagging, lemmatization, named entity
recognition (NER)~\cite{finkel2005incorporating}, temporal
tagging~\cite{chang2012sutime}, and dependency parsing~\cite{chen2014fast}.

\subsection{The MinIE-SpaTe OIE System}

From each of the resulting sentences, we extract triples using the
state-of-the-art OIE system MinIE~\cite{gashteovski2017minie}, which in turn in
based on ClausIE~\cite{del2013clausie}. MinIE \textit{minimizes} the extractions
into more compact triples by removing unnecessary words (e.g. determiners)
without damaging the semantic content of the triple and by providing
\emph{semantic annotations}. The semantic annotations move auxiliary information
from the triple (thereby simplifying it) to annotations. The semantic
annotations include polarity, modality, attribution, and quantities. Polarity
(positive or negative) indicates whether or not the triple occurred in negated
form in the input sentence. \emph{Modality} indicates whether the triple is a
certainty (CT) or merely a possibility (PS) according to the input.
\emph{Attribution} refers to the supplier of the information carried within the
triple, again according to the input sentence. Attributions have their own
polarity and modality. Finally, quantities express specific amounts.
 
Consider for example the sentence: 
\textit{``David Heyman \textcolor{forestgreen}{said} that Gilderoy Lockhart 
\textcolor{orange}{will probably} \textcolor{red}{not} be played by Kenneth Branagh.''}
MinIE extracts the following triple and annotations:
\begin{itemize}
\item[] (``Gilderoy Lockhart''; ``be played by''; ``Kenneth Branagh'')\\
  \emph{Factuality:} \textcolor{red}{negative} \textcolor{orange}{possibility} \\
	\emph{Attribution:} ``David Heyman'', (positive \textcolor{forestgreen}{certainty}) \\
\end{itemize}

Many of the sentences in Wikipedia contain some sort of temporal or spatial
reference (roughly 56\% according to a preliminary study). Since this
information is important, we modified MinIE's output to add additional semantic
annotations for space and time, thereby producing SPOTL
facts~\cite{hoffart2013yago2}. We refer to the resulting system as MinIE-SpaTe.
Generally, MinIE-SpaTe makes use of syntactic information provided in the
dependency parse as well as information provided by SUTime and the NER system.
Details as well as a discussion on the precision of the annotations can be found
in Appendix~\ref{appendix:SpaTe}.

We subsequently refer to a triple with any spatial or temporal annotation as a
\textit{spatial/temporal triple}. MinIE-SpaTe differentiates between three types
of such spatial or temporal annotations: (i) annotations on entire triples, (ii)
annotations on arguments, and (iii) spatial or temporal references. In what
follows, we briefly discuss these types for temporal annotations;
similar distinctions apply to spatial annotations.


Temporal annotations on triples provide temporal context for the entire triple.
For example, from the sentence \textit{``Bill Gates founded Microsoft
  \textcolor{blue}{in 1975}.''}, MinIE--SpaTe extracts triple \emph{(``Bill
  Gates''; ``founded''; ``Microsoft'')} with temporal annotation
\textcolor{blue}{(``in'', 1975)}. Here, ``in'' is a \textit{lexicalized
  temporal predicate} and 1975 is the \textit{core temporal expression}.
MinIE-SpaTe outputs \textit{temporal modifiers} when found; e.g. the temporal
annotation for \textit{``... at precisely 11:59 PM''} is \textit{(``at'',
  ``11:59 PM'', premod: precisely)}). We use the TIMEX3 format
\cite{sauri2006timeml} for representing the temporal information about the
triple.

Sometimes the arguments of a triple contain some temporal information that
refers to a phrase but not to the whole triple. MinIE provides temporal
annotations for arguments for this purpose. For example, from the sentence
\textit{``Isabella II opened the 17th-century Parque del Retiro.''}, MinIE-SpaTe
extracts \textit{(``Isabella II''; ``opened''; ``Parque del Retiro'')} with a
temporal annotation \textcolor{blue}{(17th-century, Parque del Retiro)} for the
object argument. Generally, the temporal annotation contains information on its
target (e.g., object), the temporal expression (17th-Century) and the head word
being modified by the temporal expression (Retiro).

Finally, some triples contain temporal references as subject or object;
MinIE-SpaTe annotates such references. For example, from the input sentence
\textit{``\textcolor{blue}{2003} was a hot year.''}, MinIE-SpaTe extracts the
triple: \textit{(``\textcolor{blue}{2003}''; ``was''; ``hot year'')}, where the
subject (\textit{\textcolor{blue}{2003}}) is annotated with a temporal
reference.

\subsection{Postprocessing} 

During postprocessing, we remove clearly wrong triples, annotate the remaining
triples with a confidence score, and ensure that Wikipedia links are not broken
up.

In particular, MinIE retains the NER types provided during preprocessing in its
extractions. We aggregated the most frequent relations per argument type and
found that many of the triples were of the form \textit{(person; ``be'';
  organization)}, \textit{(location; ``be''; organization)}, and so on. These
extractions almost always stemmed from an incorrect dependency parses obtained
from sentences containing certain conjunctions. We filtered all triples with
lemmatized relation \textit{``be''} and different NER types for subject and
object from MinIE-SpaTe's output.

In order to estimate whether or not a triple is correctly extracted, we followed
ReVerb~\cite{fader2011identifying} and trained a logistic regression classifier.
We used the labeled datasets provided by~\citet{gashteovski2017minie} as
training data. Features were constructed based on an in-depth error analysis.
The most important features were selected using chi-square relevance tests and
include features such as the clause type, whether a coordinated conjunction has
been processed, and whether or not MinIE simplified the triple. See
Appendix~\ref{appendix:conf} for a complete list of features, and
Sec.~\ref{subsec:corpus_precision} for a discussion of confidence scores and
precision in our corpora.

In a final postprocessing step, we rearranged triples such that links within the
triples are not split across its constituents. For example, the triple
\textit{(``Peter Brooke''; ``was \underline{member of}'';
  ``\underline{Parliament}'')} produced by MinIE splits up the linked phrase
``\underline{member of Parliament}'', which results an incorrect links
for the object (since it does not link to \emph{Parliament}, but to 
\emph{member of parliament}). We thus rewrite such triples to \textit{(``Peter Brooke'';
  ``was''; ``\underline{member of} \\\underline{Parliament}'')}.

\subsection{Provided Metadata}

All metadata collected in the pipeline are retained for each triple, including
provenance information, syntactic annotations, semantic annotations, and
confidence scores. A full description of the provided metadata fields can be
found in Appendix~\ref{appendix:meta-data}.


\subsection{Filtering} 

We constructed the OPIEC-Clean and OPIEC-Linked subcorpus by filtering the OPIEC
corpus. OPIEC-Clean generally only retains triples between entities or concepts,
whereas OPIEC-Linked only retains triples in which both arguments are linked.
The filtering rules are described in more details in the following
section. 


\index{\section}\section{Statistics} \label{lbl-chapter4}

Basic statistics such as corpus sizes, frequency of various semantic
annotations, and information about the length of the extracted triples of OPIEC
and its subcorpora are shown in Tab.~\ref{tab:basic_stats_for_OIE_corpora}. We
first discuss properties of the OPIEC corpus, then describe how we constructed
the OPIEC-Clean and OPIEC-Linked subcorpora, and finally provide more in-depth
statistics.

\begin{table}[t!]
  \small
	\centering
	 \begin{tabular}{lr@{~}rr@{~}rr@{~}r} \toprule
							& \multicolumn{2}{c}{OPIEC} & \multicolumn{2}{c}{OPIEC-Clean} 	& \multicolumn{2}{c}{OPIEC-Linked}    \\ \midrule
		Total triples (millions)            	& 341.0     &    & 104.0 	&  & 5.8     			&     \\ \midrule
		Triples with semantic annotations  	 & 166.3 & (49\%)& 51.46 & (49\%) & 3.37     & (58\%)     \\ 
			\hspace{2em} negative polarity   & 5.3   & (2\%) & 1.33 & (1\%)	& 0.01     & (0\%)      \\ 
			\hspace{2em} possibility modality& 13.9  & (4\%) & 3.27 & (3\%)	& 0.04     & (1\%)      \\ 
			\hspace{2em} quantities          & 59.4  & (17\%)& 15.91 & (15\%)& 0.45     & (8\%)      \\ 
			\hspace{2em} attribution         & 6.4   & (2\%) & 1.44 & (1\%)	& 0.01     & (0\%)      \\ 
			\hspace{2em} time                & 65.3  & (19\%)& 19.66 & (19\%)& 0.58     & (1\%)      \\ 
			\hspace{2em} space               & 61.5  & (18\%)& 22.11 & (21\%)& 2.64     & (45\%)     \\ 
			\hspace{2em} space OR time       & 111.3 & (33\%)& 37.22 & (36\%)& 3.01     & (52\%)     \\ 
			\hspace{2em} space AND time      & 15.4  & (5\%) & 4.54 & (4\%)	& 0.20     & (4\%)      \\ \midrule
		Triple length in tokens ($\mu \pm \sigma$)& \multicolumn{2}{r}{$7.66 \pm 4.25$} & \multicolumn{2}{r}{$6.06 \pm 2.82$} & \multicolumn{2}{r}{$6.45 \pm 2.65$} \\ 
		\hspace{2em} subject ($\mu \pm \sigma$)   & \multicolumn{2}{r}{$2.12 \pm 2.12$} & \multicolumn{2}{r}{$1.48 \pm 0.79$} & \multicolumn{2}{r}{$1.92 \pm 0.94$} \\ 
		\hspace{2em} relation ($\mu \pm \sigma$)  & \multicolumn{2}{r}{$3.01 \pm 2.47$} & \multicolumn{2}{r}{$3.10 \pm 2.56$} & \multicolumn{2}{r}{$2.77 \pm 2.14$} \\ 
		\hspace{2em} object ($\mu \pm \sigma$)    & \multicolumn{2}{r}{$2.52 \pm 2.69$} & \multicolumn{2}{r}{$1.48 \pm 0.79$} & \multicolumn{2}{r}{$1.76 \pm 0.94$} \\ \midrule
		Confidence score ($\mu \pm \sigma$) & \multicolumn{2}{r}{$0.53 \pm 0.23$} & \multicolumn{2}{r}{$0.59 \pm 0.23$} & \multicolumn{2}{r}{$0.61 \pm 0.26$} \\ \bottomrule
 	\end{tabular}
	\caption{Statistics for different OPIEC corpora. 
	All frequencies are in millions. We count triples with
    annotations (not annotations directly). Percentages refer to the respective
    subcorpus.}
	\label{tab:basic_stats_for_OIE_corpora}
\end{table}

\subsection{The OPIEC Corpus}

The OPIEC corpus contains all extractions produced by MinIE-SpaTe. We analyzed
these extractions and found that a substantial part of the triples are more
difficult to handle by downstream applications. We briefly summarize the most
prevalent cases of such triples; all these triples are filtered out in
OPIEC-Clean.

First of all, a large part of the triples are under-specific in that additional
context information from the extraction source is required to obtain a coherent
piece of information. By far the main reason for under-specificity is lack of
coreference information. In particular, 22\% of the arguments in OPIEC are
personal pronouns, such as in the triple \textit{(``He''; ``founded'';
  ``Microsoft'')}. Such triples are under-specific because provenance
information is needed to resolve what ``He'' refers to. Similarly, about 1\% of
the triples have determiners as arguments (e.g. \textit{(``This''; ``lead to'';
  ``controversy'')}), and 0.2\% Wh-pronouns (e.g. \textit{(``what''; ``are known
  as''; ``altered states of consciousness'')}). Coreference resolution in itself
is a difficult problem, but the large fraction of such triples shows that
coreference resolution is important to further boost the recall of OIE systems.

Another problem for OIE systems are entity mentions---most notably for works of
art---that constitute clauses. For example, the musical \textit{``Zip Goes a
  Million''} may be interpreted as a clause, leading to the incorrect extraction
\textit{(``\underline{Zip}''; ``\underline{Goes}''; ``\underline{a Million}'')}.
A preliminary study showed that almost 30\% of all the OPIEC triples containing
the same recognized named entity in both subject and object were of such a type.
These cases constitute around 1\% of OPIEC.

Finally, a substantial fraction of the triples in OPIEC has complicated
expressions in its arguments. Consider for example the sentence \textit{``John
  Smith learned a great deal of details about the U.S. Constitution.''}.
MinIE extracts the triple
\textit{(``John Smith''; ``learned''; ``great deal of details about U.S.
  Constitution'')}, which has a complicated object and is thus difficult to
handle. A minimized variant such as \textit{(``John Smith''; ``learned about'';
  ``U.S. Constitution'')} looses some information, but it expresses the main
intent in a simpler way.

The above difficulties make the OPIEC corpus a helpful resource for research on
improving or reasoning with complex OIE extractions rather than for downstream
tasks.

\subsection{The OPIEC-Clean Corpus}

The OPIEC-Clean corpus is obtained from OPIEC by simply removing underspecified
and complex triples. In particular, we consider a triple \emph{clean} if the
following conditions are met: (i) each argument is either linked, an entity
recognized by the NER tagger, or matches a Wikipedia page title, (ii) links or
recognized named entities are not split up across constituents, and (iii) the
triple has a non-empty object.

Conditions (i) and (ii) rule out the complex cases mentioned in the previous
section. Note that we ignore quantities (but no other modifiers) when checking
condition (i). For example, the triple \textit{(``$Q_1$ \underline{electric
  locomotives}''; ``were ordered from''; ``\underline{Alsthom}'')} with
\textit{$Q_1=$``Three''} is considered clean; here \textit{``electric locomotives''}
holds a link to \emph{TCDD E4000} and \textit{``Alsthom''} holds a link to
\emph{Alsthom}.

MinIE is a clause-based OIE system and can produce extractions for
so-called SV clauses: these extractions consists of only a subject and a
relation, but no object. 3.5\% of the triples in OPIEC are of such type. An
example is the triple \textit{(``Civil War''; ``have escalated''; ``'')}.
Although such extractions may contain useful information, we exclude them via
condition (iii) to make the OPIEC-Clean corpus uniform.

Roughly 30\% of the triples (104M) in OPIEC are clean according to the above
constraints. Table~\ref{tab:basic_stats_for_OIE_corpora} shows that clean
triples are generally shorter on average and tend to have a higher confidence
score than the full set of triples in OPIEC. The OPIEC-Clean corpus is easier to
work with than the full OPIEC corpus; it is targeted towards both downstream
applications and research in automated knowledge base construction.

\subsection{The OPIEC-Linked Corpus}

The OPIEC-Linked corpus contains only those triples from OPIEC-Clean in which
both arguments are linked. Although the corpus is much smaller than OPIEC-Clean
(5.8M triples, i.e., roughly 5.5\% of OPIEC-Clean), it is the largest corpus to
date with golden disambiguation links for the arguments. We use the corpus
mainly to compare OIE extractions with the information present in the DBpedia
and YAGO knowledge bases; see Sec.~\ref{sec:analysis}.

\subsection{Semantic Annotations}

About 49\% of all triples in OPIEC contain some sort of semantic annotation
(cf.~Tab.~\ref{tab:basic_stats_for_OIE_corpora}); in OPIEC-Linked, the fraction
increases to 58\%. Most of the semantic annotations referred to quantities, space
or time; these annotations provide important context for the extractions. There
is a significantly smaller amount of negative polarity and possibility modality
annotations. One reason for the lack of such annotations may be in the nature of
the Wikipedia articles, which aim to contain encyclopedic, factual statements and
are thus more rarely negated or hedged.

The distribution of annotations is similar for OPIEC and OPIEC-Clean, but
differs significantly for OPIEC-Linked. In particular, we observe a drop in
quantity annotations in OPIEC-Linked because most of the linked phrases do not
contain quantities. The fraction of spatial triples in OPIEC-Linked is also much
higher than the rest of the corpora. The reason is because Wikipedia contains many
pages for locations, which are also linked and contain factual knowledge, 
thus resulting in many triples with spatial reference
(e.g. the triple \textit{(``\underline{Camborne School of Mines}''; ``was relocated to''; 
``\underline{Penryn}'')}).

\subsection{NER Types and Frequent Relations}

\begin{figure}
    \centering
    \includegraphics[width=0.6\textwidth]{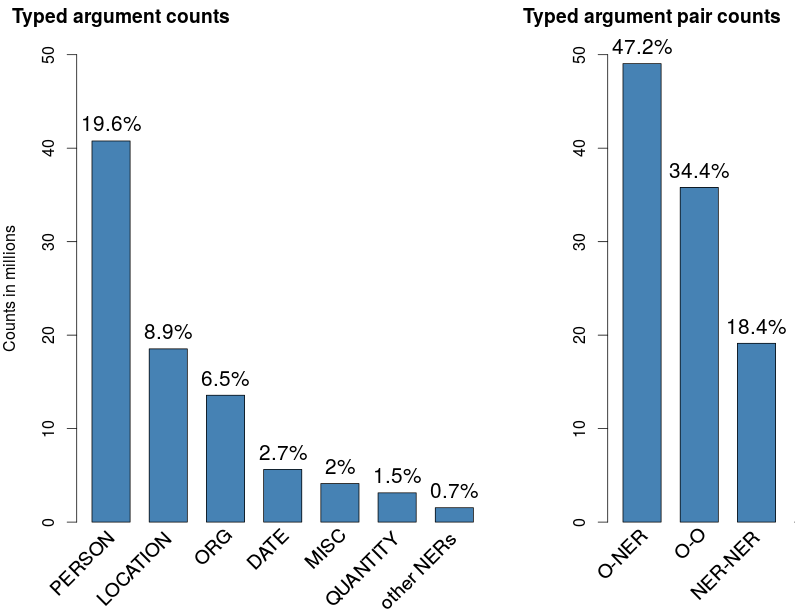}
    \caption{Distribution of NER types for arguments and argument pairs in OPIEC-Clean. Here ``O'' refers to arguments that are not recognized as a named entity.}
    \label{fig:ner-stats}
\end{figure}

For OPIEC-Clean, Fig.~\ref{fig:ner-stats} shows the fraction of arguments and
argument pairs that are recognized as named entities by the NER tagger, along
with the NER type distribution of the arguments.

Out of the around 208M arguments, roughly 42\% are recognized named entities.
The most frequent NER type is \emph{person}, followed by \emph{location} and
\emph{organization}. The remaining NER types are not that frequent (less than
3\% each). On the other hand, 58\% of the arguments are not typed. These are
mostly concepts (more precisely, strings that match Wikipedia pages not
referring to an entity) and are thus not recognized by the NER system. 
The top-10 most frequent arguments which are not typed are the words \textit{film,
population, city, village, father, song, company, town, album} and \textit{time},
with frequencies varying between 408k and 616k.

Fig.~\ref{fig:ner-stats} also reports the fraction of triples in which none, one,
or both arguments are recognized as a named entity. We found that 18\% of the
triples (19M) in OPIEC-Clean have two typed arguments, and around 66\% of the
triples (68M) have at least one typed argument. Thus the majority of the triples
involves named entities. 34\% of the triples do not have recognized named entity
arguments.


Tab.~\ref{tab:rels_between_entity_pairs} shows the most frequent open
relations between arguments recognized as persons and/or locations (which in
turn are the top-3 most frequent argument type pairs). We will analyze some of
the open relations in more detail in Sec.~\ref{sec:analysis}. For now, note
that the most frequent relation between persons is \textit{``have''}, which is highly
polysemous. Other relations, such as ``\textit{marry}'' and \textit{``be son
  of''}, are much less ambiguous. We provide all open relations between
recognized argument types as well as their frequencies with the OPIEC-Clean
corpus.


\begin{table}
	\small
	\centering
	\begin{tabular}{rrrrrr} 				\toprule
	 	\multicolumn{2}{r}{PERSON-PERSON} & \multicolumn{2}{r}{LOCATION-LOCATION}	& \multicolumn{2}{r}{PERSON-LOCATION} \\ \midrule 	
		\textit{"have"} &(130,019) 				& \textit{"be in"} 						&(2,126,562)& \textit{"be bear in"} 	& (203,091) \\			
		\textit{"marry"} &(49,405) 				& \textit{"have"} 							&(40,298)	& \textit{"die in"} 		& (37,952) 	\\			
		\textit{"be son of"} &(40,265)				& \textit{"be village in adminis-}		& 			& \textit{"return to"} 	& (36,702) 	\\			
						&				& \textit{trative district of"}			& (9,130)	&  				&			\\			
		\textit{"be daughter of"} & (37,089) 		& \textit{"be north of"} 					&(3,816) 	& \textit{"move to"} 		& (36,072)  \\			
		\textit{"be bear to"} &(29,043) 			& \textit{"be suburb of"} 					&(3,291) 	& \textit{"be in"} 		& (25,847) \\			
		\textit{"be know as"} &(25,607)			& \textit{"be west of"} 					& (3,238) 	& \textit{"live in"} 		& (22,399) \\			
		\textit{"defeat"} &(22,151) 				& \textit{"be part of"} 					&(3,188) 	& \textit{"grow up in"} 	& (17,571) \\ \bottomrule 
	\end{tabular}
	\caption{Most frequent open relations between persons and locations (as
    recognized by the NER tagger) in OPIEC-Clean}
	\label{tab:rels_between_entity_pairs}
\end{table}


\subsection{Precision and Confidence Score}
\label{subsec:corpus_precision}

\begin{figure}
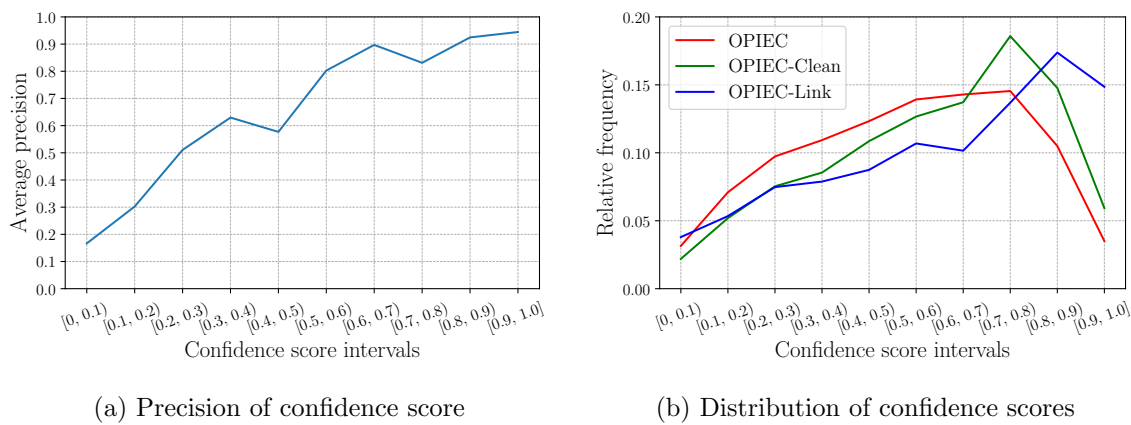

    \centering
    \begin{subfigure}[t]{0.5\textwidth}
        \centering
        \scalebox{0.5}{\input{fig3a_2.pgf}}
        \caption{Precision of confidence score}
        \label{fig:distr-corr-a}
    \end{subfigure}%
    ~
    \begin{subfigure}[t]{0.5\textwidth}
        \centering
        \scalebox{0.5}{\input{fig3b_2.pgf}}
        \caption{Distribution of confidence scores}
        \label{fig:distr-corr-b}
    \end{subfigure}
    \caption{Precision and distribution of confidence scores}
    \label{fig:distr-corr}
\end{figure}

Each triple in the OPIEC corpora is annotated with a confidence score for
correct extraction. To evaluate the accuracy of the confidence score, we took an
independent random sample of triples (500 in total) from OPIEC and manually
evaluated the correctness of the triples following the procedure
of~\citet{gashteovski2017minie}.\footnote{The annotation guidelines are provided
  at
  \url{http://www.uni-mannheim.de/media/Einrichtungen/dws/Files\_Research/Software/MinIE/minie-labeling-guide.pdf}}
We found that 355 from the 500 triples (71\%) were correctly extracted. Next, we
bucketized the triples by confidence score into ten equi-width intervals and
calculated the precision within each interval; see Fig. \ref{fig:distr-corr-a}.
We found that the confidence score is highly correlated to precision (Pearson
correlation of $r=0.95$) and thus provides useful information.




The distribution in confidence scores across the various corpora is shown in
Fig.~\ref{fig:distr-corr-b}. We found that OPIEC-Clean and, in particular,
OPIEC-Link contain a larger fraction of high-confidence triples than the raw
OPIEC corpus: these corpora are cleaner in that more potentially erroneous
extractions are filtered out. Although triples with lower confidence score tend
to be more inaccurate, they may still provide suitable information. We thus
included these triples into our corpora; the confidence scores allow downstream
applications to handle these triples as appropriate.



\section{Analysis} 
\label{sec:analysis}

In this section, we compare the information present in the OIE triples in OPIEC
with the information present in the DBpedia~\cite{auer2007dbpedia} and
YAGO~\cite{hoffart2013yago2} knowledge bases. Since all resources extract
information from Wikipedia--- OPIEC from the text and DBpedia as well as YAGO from
semi-structured parts of Wikipedia---, we wanted to understand whether and to what
extent they are complementary. Generally, the disambiguation of OIE triples
w.r.t.~a given knowledge base is in itself a difficult problem. We avoid this
problem here by (1) restricting ourselves to the OPIEC-Linked corpus (for which
we have golden entity links) and (2) focusing on statistics that do not
require a full disambiguation of the open relations but are nevertheless
insightful.

\subsection{Alignment With Knowledge Bases}

To align the OIE tripes from OPIEC-Linked to YAGO or DBpedia, we make use of the
distant supervision assumption. For each open triple $(s,r_{\text{open}},o)$
from OPIEC-Linked, we search the KB for any triple of form $(s,r_{\text{KB}},o)$
or $(o,r_{\text{KB}},s)$. Here $s$ and $o$ refer to disambiguated entities,
whereas $r_{\text{open}}$ refers to an open relation and $r_{\text{KB}}$ to a KB
relation. If such a triple exists, we say that triple $(s,r_{\text{open}},o)$
has a \emph{KB hit}, and that $r_{\text{open}}$ \emph{is aligned with}
$r_{\text{KB}}$. In fact, if $r_{\text{open}}$ (e.g., \textit{``was born in''})
is a mention of KB relation $r_{\text{KB}}$ (e.g., \textit{birthPlace}), and
subject and object are not swapped, then both triples express the same
information. Otherwise, the open triple and the KB triple express different
information (e.g., \textit{deathPlace}) or inverse relations (e.g.,
\textit{isBirthplaceOf}). We can thus think of the number of KB hits as an
optimistic measure of the number of open triples that are represented in the KB
(with caveats, see below): the KB contains some relation between the
corresponding entities, although not necessarily the one being mentioned.

We observed that 29.7\% of the OIE triples in OPIEC-Linked have a KB hit in
either DBpedia or YAGO. More specifically, 25.5\% of the triples have a KB hit
in DBpedia, 20.8\% in YAGO, and 16.6\% in both DBpedia and YAGO. Most of these
triples have exactly one hit in the corresponding KB. Consequently, 70.3\% of
the linked triples do not have a KB hit; we analyze these triples below.





Tab.~\ref{tab:rel_clusters} shows the most frequent open relations aligned to
the DBpedia relations \emph{location}, \emph{associatedMusicalArtist}, and
\emph{spouse}. The frequencies correspond to the number of OIE triples that (1)
have the specified open relation (e.g., \textit{``be wife of''}) and (2) have a KB hit
with the specified KB relation (e.g., \textit{spouse}). There is clearly no 1:1
correspondence between open relations and KB relations. On the one hand, open
relations can be highly ambiguous (e.g., \textit{``be''} has hits to \emph{location} and
\emph{associatedMusicalArtits}). On the other hand, open relations can also
be more specific than KB relations (e.g., \textit{``be guitarist of''} is more specific
than \emph{associatedMusicalArtist}) or semantically different (e.g., \textit{``be widow of''} 
and \emph{spouse}) than the KB relations they align to.

\begin{table}
	\small
	\centering
	\begin{tabular}{rrrrrr} \toprule
		\multicolumn{2}{c}{\hspace{1cm}location}	&\multicolumn{2}{c}{associatedMusicalArtist} 	&	\multicolumn{2}{c}{spouse} 	\\ \midrule
		\textit{``be in''} 		& (43,842)	& \textit{``be''}   		&(6,273)	& \textit{``be wife of''} 	&(1,965)\\ 
		\textit{``have''} 		& (3,175)	& \textit{``have''} 		&(3,600)	& \textit{``be''}		&(1,308)\\ 
		\textit{``be''}			& (1,901)	& \textit{``be member of''}	&(740)		& \textit{``marry''}		&(702)	\\ 
		\textit{``be at''} 		& (1,109)	& \textit{``be guitarist of''} 	&(703) 		& \textit{``be widow of''} 	&(479)	\\ 
		\textit{``be of''} 		& (706)		& \textit{``be drummer of''}	&(458) 		& \textit{``have''} 		&(298)	\\ 
		\textit{``be historic home}	& (491)		& \textit{``be feature''}	&(416)		& \textit{``be husband of''} 	&(284) 	\\	
		\textit{located at''} 	 	&   		& 				&		& 				&	\\ \bottomrule
 	\end{tabular}
	\caption{The most frequent open relations aligned to the DBpedia relations
    \emph{location}, \emph{associatedMusicalArtist}, and \emph{spouse} in
    OPEIC-Linked}
	\label{tab:rel_clusters}
\end{table}


To gain more insight into the type of triples contained in OPIEC-Clean, we
selected the top-100 most frequent open relations for further analysis. These
relations constitute roughly 38\% of the OPIEC-Clean corpus, relation
frequencies are thus highly skewed. We then used OPIEC-Linked as a proxy for the
number of DBpedia hits of these relations. The results are summarized in
Tab.~\ref{tab:rel_ratios} as well as in 
Appendix~\ref{appendix:kb-hits}.
The open relation \textit{``have''}, for example, is
aligned to 330 distinct DBpedia relations, the most frequent ones being
\textit{author, director,} and \textit{writer}. Generally, the fraction of KB
hits (from OPIEC-Linked) is quite low, averaging at 16.8\% for the top-100
relations. This indicates that there is a substantial amount of information
present in OIE triples that is not present in KBs. Moreover, about 42 distinct
KB relations align on average with each open relation, 
which again indicates that open
relations should not be directly mapped to KB relations.

\begin{table}[p!]
	\footnotesize
	\centering
	\begin{tabular}{l@{\hspace{-.2cm}}rrr@{~}rcl@{\hspace{-.5cm}}r} \toprule
		Open                & Frequency in	& Frequency in 	& \multicolumn2c{\# KB hits} 	& \# distinct& \multicolumn2c{Top-3 aligned DBpedia rel.}      \\ 
		relation            & OPIEC-Clean	& OPIEC-Link   	&                            	&         & KB rel.s 	& \multicolumn2c{and hit frequency} \\ \midrule
	      \textit{``be''}       & 21,911,174   	& 1,475,332	& 173,107                    	& (11.7\%)& 410 	& type & 72,077                	\\ 
				    &              	&              	&                            	&         &         	& occupation & 12,508          	\\ 
				    &              	&              	&                            	&         &         	& isPartOf & 8,012             	\\ \midrule
	      \textit{``have''}     & 6,369,086    	&   216,332    	& 137,865                    	& (63.7\%)& 330 	& author & 14,056              	\\  
				    &              	&              	&                            	&         &         	& director & 10,416             \\ 
				    &              	&              	&                            	&         &         	& writer & 9,765               	\\ \midrule
	      \textit{``be in''}    & 3,219,301    	& 1,150,667	& 804,378                    	& (69.9\%)& 225 	& country & 287,557            	\\ 
				    &              	&              	&                            	&         &         	& isPartOf & 222,175           	\\ 
				    &              	&              	&                            	&         &         	& state & 64,675               	\\ \midrule
	    \textit{``include''}    &   487,899      	&    14,746   	&   1,573                      	& (10.7\%)& 128 	& type & 380                   	\\
				    &              	&              	&                            	&         &         	& associatedBand & 83          	\\ 
				    &              	&              	&                            	&         &         	& associatedMusicalArtist & 83 	\\ \midrule
	    \textit{``be bear in''} &   289,947      	&     7,138    	&    1,477                     	& (20.7\%)& 30  	& birthPlace & 1,147           	\\ 
				    &              	&              	&                            	&         &         	& isPartOf & 73                	\\ 
				    &              	&              	&                      		&         &         	& deathPlace & 62              	\\ \midrule
		    \textit{``win''}&   236,169      	&     8,819    	&     910		   	& (10.3\%)& 54       	& award & 299 		       	\\ 
				    &         		&         	&       			&         &           	& race & 210                   	\\
				    &         		&         	&       			&         &           	& team & 50 		       	\\ \midrule
	     \textit{``be know as''}&   215,809    	&     7,993     &     675     			& (8.4\%) & 123       	& location & 46 		\\ 
				    &         		&         	&       			&         &           	& associatedBand & 42 		\\ 
				    &         		&         	&       			&         &           	& associatedMusicalArtist & 42 	\\ \midrule
		 \textit{``become''}&   213,807       	&     5,123     &     393     			& (7.7\%) & 90        	& successor & 63 		\\ 
				    &         		&         	&       			&         &           	& associatedBand & 33 		\\
				    &         		&         	&       			&         &           	& associatedMusicalArtist & 33 	\\ \midrule
		\textit{``have be''}&   191,140       	&     1,855     &     101      			& (5.4\%) & 32        	& type & 12 			\\  
				    &         		&         	&       			&         &           	& position & 9 			\\ 
				    &         		&         	&       			&         &           	& leader & 7 			\\ \midrule
		   \textit{``play''}&   163,643       	&     4,842     &     835      			& (17.2\%)& 54      	& portrayer & 367 		\\  
				    &         		&         	&       			&         &           	& author & 101 			\\ 
				    &         		&         	&       			&         &           	& instrument & 76 		\\ \midrule
		\textit{``be know''}&   157,751       	&       351     &      51      			& (14.5\%)& 15        	& occupation & 20 		\\ 
				    &         		&         	&       			&         &           	& family & 12 			\\ 
				    &         		&         	&       			&         &           	& country & 3 			\\ \midrule
		 \textit{``die in''}&   146,681       	&       638     &     127      			& (19.9\%)& 20        	& deathPlace & 71 		\\ 
				    &         		&         	&       			&         &           	& battle & 13 			\\
				    &         		&         	&       			&         &           	& commander & 9 		\\ \midrule
		  \textit{``join''} &   134,159       	&     2,656     &     903      			& (34.0\%)& 65        	& team & 301 			\\ 
				    &         		&         	&       			&         &           	& associatedBand & 105 		\\
				    &         		&         	&       			&         &           	& associatedMusicalArtist & 105 \\ \midrule
 	\end{tabular}
	\caption{The most frequent open relations in OPIEC-Clean, along with DBpedia
    alignment information from OPIEC-Link (continued in Tab. 6, Appendix~\ref{appendix:kb-hits})}
	\label{tab:rel_ratios}
\end{table}

By far the most frequent open relations in OPIEC-Clean are
\textit{``be''} and \textit{``have''}, which constitute 21.1\% and 6.1\% of all
the triples, respectively. These open relations are also the most ambiguous
ones in that they are aligned with 410 and 330 different DBpedia relations,
respectively. Here the open relations are far more ``generic'' than the KB
relations that they are aligned to. This is illustrated in the following
examples:
\vspace{-1em}
\begin{center}
  \begin{tabular}{lcl}
	\textit{(``\underline{Claudia Hiersche}''; ``be''; ``\underline{actress}'')} & $\xrightarrow{DBpedia}$ & (Claudia\_Hiersche; occupation; Actress) \\
	\textit{(``\underline{Cole Porter}''; ``have''; ``\underline{Can-Can}'' )} & $\xrightarrow{DBpedia}$ & (Can-Can\_(musical); musicBy; Cole\_Porter) \\
												 &  $\xrightarrow{DBpedia}$ & (Can-Can\_(musical); lyrics; Cole\_Porter) \\
	\textit{(``\underline{Oddbins}''; ``have''; ``\underline{Wine}'')} & $\xrightarrow{DBpedia}$ & (Oddbins; product; Wine)
  \end{tabular}
\end{center}
Note that in these cases, ``have'' refers to the possessive (e.g., Odbbins'
Wine).

\subsection{Spatio-Temporal Facts} \label{subsection:spotl}

We also investigated to what extent the space and time annotations in OIE
triples relate to corresponding space and time annotations in YAGO.
In particular, YAGO provides:
\begin{itemize}
\item \emph{YAGO date facts}, which have entities as subjects and dates as
  objects: e.g., (Keith\_Joseph, wasBornOnDate, 1918-01-17).
\item \emph{YAGO meta-facts}, which are spatial or temporal information about
  other YAGO facts: e.g., (Steven\_Lennon, playsFor, Sandnes\_Ulf) has meta-fact
  (occursUntil, 2014).
\end{itemize}
Note that date facts roughly correspond to temporal reference annotations in
OPIEC, wheres meta-facts correspond to spatial or temporal triple annotations.


To compare OPIEC with YAGO date facts, we selected all triples with (i) a
disambiguated subject and (ii) an object that is annotated as date from OPIEC.
There are 645,525 such triples. As before, we align these triples to YAGO using
an optimistic notion of a KB hit. In particular, a \emph{KB date hit} for an OIE
date fact $(s, r_{\text{open}}, d_{\text{open}})$ is any KB fact of form $(s,
r_{\text{KB}}, d_{\text{KB}})$, i.e., we require that there is temporal
information but ignore whether or not it matches. We use this optimistic notion
of KB date hit to avoid disambiguating the open relation or date. Even with
this optimistic notion, we observed that only 36,262 (5.6\%) of the OIE date facts
have a KB date hit in the YAGO date facts.

We also compared the spatial and temporal annotations of OPIEC-Linked with the
YAGO meta-facts. We found that roughly 13,203 OPIEC-Linked triples have a KB hit
with a YAGO triple that also has an associated a meta-fact. Out of these linked
OIE triples, 2,613 are temporal and 2,629 are spatial. 

To provide further insight, we analyzed the spatial-temporal annotations of
OPIEC more closely. We identified two major reasons why spatio-temporal
information is not found in KBs: (i) the
information in missing from the KB, and (ii) the information is available only
indirectly. For an example of missing information, consider the OPIEC-Linked
triple \textit{(``Iain Duncan Smith'', ``is leader of'', ``Conservative Party'')} with
temporal annotations \textit{(pred=``from'', 2001)} and \textit{(pred=``to'', 2003)}. YAGO
contains the KB hit (Iain\_Duncan\_Smith; isAffiliatedTo;
Conservative\_Party\_(UK)). Note that the YAGO relation is less specific than
the open relation, and that no temporal information is present. As another
example, consider the OIE triple \textit{("Neue Nationalgalerie"; "be built by";
  "Ludwig Mises van der Rohe")} 
  with temporal
annotation \textit{(pred=``in'', 1968)}. Again, the YAGO hit (Neue\_Nationalgalerie;
linksTo; Ludwig\_Mies\_van\_der\_Rohe) is less specific and lacks temporal
information. YAGO does contain the triple (Neue\_Nationalgalerie; hasLongitude;
13.37) with temporal meta-fact 1968-01-01. Here the temporal information is
present in the KB, but only indirectly and for a different relation.

Generally, the low number of KB hits indicates that a wealth of additional
spatial and/or temporal information is present in OIE data, and that the
spatial/temporal annotations provided in OPIEC are potentially very valuable for
automated KB completion tasks.

\subsection{Non-Aligned OIE Triples}


We found that more than half of the triples in OPIEC-Linked that do not have a
KB hit refer to one of the top-100 most frequent relations in OPIEC-Clean. Since
OPIEC-Clean is much larger than OPIEC-Linked, this indicates that it contains
many facts not present in DBpedia. 
Naturally, not all these facts are correct, though, and disambiguation is a
major challenge. In particular, we took a random sample of 100 non-aligned
triples from OPIEC-Linked and manually labeled each triple as \textit{correctly
  extracted} or \textit{incorrectly extracted}. 60\% of the triples were
considered to be correctly extracted. In another sample of 100 high-confidence
triples (score $>0.5$), 80\% were correctly extracted. This shows clearly the
potential and the challenges of harnessing the knowledge contained within the
OIE triples.

\section{Conclusions}

We created OPIEC, a large open information extraction corpus extracted from
Wikipedia. OPIEC consists of hundreds of millions of triples, along with rich
metadata such as provenance information, syntactic annotations, semantic
annotations, and confidence scores. We reported on a data profiling study of the
OPIEC corpus as well as subcorpora. In particular, we analyzed to what extent
OPIEC overlaps with the DBpedia and YAGO knowledge bases. Our study indicates
that most open facts do not have counterparts in the KB such that OIE corpora
contain complementary information. For the information that overlaps, open
relation are often more specific, more generic, or simply correlated to KB
relations (instead of semantically equivalent). We hope that the OPIEC corpus,
its subcorpora, derived statistics, as well as the codebase used to create the
corpus are a valuable resource for automated KB construction and downstream
applications.

\bibliography{akbc}

\begin{thebibliography}{35}
\providecommand{\natexlab}[1]{#1}
\providecommand{\url}[1]{\texttt{#1}}
\expandafter\ifx\csname urlstyle\endcsname\relax
  \providecommand{\doi}[1]{doi: #1}\else
  \providecommand{\doi}{doi: \begingroup \urlstyle{rm}\Url}\fi

\bibitem[Auer et~al.(2007)Auer, Bizer, Kobilarov, Lehmann, Cyganiak, and
  Ives]{auer2007dbpedia}
S{\"o}ren Auer, Christian Bizer, Georgi Kobilarov, Jens Lehmann, Richard
  Cyganiak, and Zachary Ives.
\newblock D{B}pedia: {A} {N}ucleus for a {W}eb of {O}pen {D}ata.
\newblock In \emph{The Semantic Web}, pages 722--735. 2007.

\bibitem[Balasubramanian et~al.(2012)Balasubramanian, Soderland, Mausam, and
  Etzioni]{balasubramanian2012rel}
Niranjan Balasubramanian, Stephen Soderland, Mausam, and Oren Etzioni.
\newblock Rel-grams: {A} {P}robabilistic {M}odel of {R}elations in {T}ext.
\newblock In \emph{Proc. of the Joint Workshop on Automatic Knowledge Base
  Construction and Web-scale Knowledge Extraction (AKBC-WEKEX@NAACL-HLT)},
  pages 101--105, 2012.

\bibitem[Balasubramanian et~al.(2013)Balasubramanian, Soderland, Mausam, and
  Etzioni]{balasubramanian2013generating}
Niranjan Balasubramanian, Stephen Soderland, Mausam, and Oren Etzioni.
\newblock Generating {C}oherent {E}vent {S}chemas at {S}cale.
\newblock In \emph{Proc. of the Conference on Empirical Methods in Natural
  Language Processing (EMNLP)}, pages 1721--1731, 2013.

\bibitem[Banko et~al.(2007)Banko, Cafarella, Soderland, Broadhead, and
  Etzioni]{banko2007open}
Michele Banko, Michael~J. Cafarella, Stephen Soderland, Matthew Broadhead, and
  Oren Etzioni.
\newblock Open {I}nformation {E}xtraction from the {W}eb.
\newblock In \emph{Proc. of the International Joint Conferences on Artificial
  Intelligence (IJCAI)}, volume~7, pages 2670--2676, 2007.

\bibitem[Bizer et~al.(2009)Bizer, Lehmann, Kobilarov, Auer, Becker, Cyganiak,
  and Hellmann]{bizer2009dbpedia}
Christian Bizer, Jens Lehmann, Georgi Kobilarov, S{\"o}ren Auer, Christian
  Becker, Richard Cyganiak, and Sebastian Hellmann.
\newblock D{B}pedia - {A} {C}rystallization {P}oint for the {W}eb of {D}ata.
\newblock \emph{Web Semantics: Science, Services and Agents on the World Wide
  Web}, 7\penalty0 (3):\penalty0 154--165, 2009.

\bibitem[Chang and Manning(2012)]{chang2012sutime}
Angel~X. Chang and Christopher~D. Manning.
\newblock S{UT}ime: {A} {L}ibrary for {R}ecognizing and {N}ormalizing {T}ime
  {E}xpressions.
\newblock In \emph{Proc. of the Conference on Language Resources and Evaluation
  (LREC)}, pages 3735--3740, 2012.

\bibitem[Chen and Manning(2014)]{chen2014fast}
Danqi Chen and Christopher Manning.
\newblock A {F}ast and {A}ccurate {D}ependency {P}arser using {N}eural
  {N}etworks.
\newblock In \emph{Proc. of the Conference on Empirical Methods in Natural
  Language Processing (EMNLNP)}, pages 740--750, 2014.

\bibitem[Del~Corro and Gemulla(2013)]{del2013clausie}
Luciano Del~Corro and Rainer Gemulla.
\newblock Claus{IE}: {C}lause-{B}ased {O}pen {I}nformation {E}xtraction.
\newblock In \emph{Proc. of the Conference on World Wide Web (WWW)}, pages
  355--366, 2013.

\bibitem[Delli~Bovi et~al.(2015{\natexlab{a}})Delli~Bovi, Espinosa-Anke, and
  Navigli]{delli2015knowledge}
Claudio Delli~Bovi, Luis Espinosa-Anke, and Roberto Navigli.
\newblock Knowledge {B}ase {U}nification via {S}ense {E}mbeddings and
  {D}isambiguation.
\newblock In \emph{Proc. of the Conference on Empirical Methods in Natural
  Language (EMNLP)}, pages 726--36, 2015{\natexlab{a}}.

\bibitem[Delli~Bovi et~al.(2015{\natexlab{b}})Delli~Bovi, Telesca, and
  Navigli]{bovi2015large}
Claudio Delli~Bovi, Luca Telesca, and Roberto Navigli.
\newblock Large-{S}cale {I}nformation {E}xtraction from {T}extual {D}efinitions
  through {D}eep {S}yntactic and {S}emantic {A}nalysis.
\newblock \emph{Transactions of the Association for Computational Linguistics
  (TACL)}, 3:\penalty0 529--543, 2015{\natexlab{b}}.

\bibitem[Dong et~al.(2014)Dong, Gabrilovich, Heitz, Horn, Lao, Murphy,
  Strohmann, Sun, and Zhang]{dong2014knowledge}
Xin Dong, Evgeniy Gabrilovich, Geremy Heitz, Wilko Horn, Ni~Lao, Kevin Murphy,
  Thomas Strohmann, Shaohua Sun, and Wei Zhang.
\newblock Knowledge {V}ault: {A} {W}eb-{S}cale {A}pproach to {P}robabilistic
  {K}nowledge {F}usion.
\newblock In \emph{Proc. of the ACM SIGKDD Conference on Knowledge Discovery
  and Data Mining (KDD)}, pages 601--610, 2014.

\bibitem[Etzioni et~al.(2008)Etzioni, Banko, Soderland, and
  Weld]{etzioni2008open}
Oren Etzioni, Michele Banko, Stephen Soderland, and Daniel~S Weld.
\newblock Open {I}nformation {E}xtraction from the {W}eb.
\newblock \emph{Communications of the ACM}, 51\penalty0 (12):\penalty0 68--74,
  2008.

\bibitem[Fader et~al.(2011)Fader, Soderland, and Etzioni]{fader2011identifying}
Anthony Fader, Stephen Soderland, and Oren Etzioni.
\newblock Identifying {R}elations for {O}pen {I}nformation {E}xtraction.
\newblock In \emph{Proc. of the Conference on Empirical Methods in Natural
  Language Processing (EMNLP)}, pages 1535--1545, 2011.

\bibitem[Fader et~al.(2013)Fader, Zettlemoyer, and
  Etzioni]{fader2013paraphrase}
Anthony Fader, Luke Zettlemoyer, and Oren Etzioni.
\newblock Paraphrase-{D}riven {L}earning for {O}pen {Q}uestion {A}nswering.
\newblock In \emph{Proc. of the Annual Meeting of the Association for
  Computational Linguistics (ACL)}, volume~1, pages 1608--1618, 2013.

\bibitem[Finkel et~al.(2005)Finkel, Grenager, and
  Manning]{finkel2005incorporating}
Jenny~Rose Finkel, Trond Grenager, and Christopher Manning.
\newblock Incorporating {N}on-local {I}nformation into {I}nformation
  {E}xtraction {S}ystems by {G}ibbs {S}ampling.
\newblock In \emph{Proc. of the Annual Meeting on Association for Computational
  Linguistics (ACL)}, pages 363--370, 2005.

\bibitem[Gashteovski et~al.(2017)Gashteovski, Gemulla, and
  Del~Corro]{gashteovski2017minie}
Kiril Gashteovski, Rainer Gemulla, and Luciano Del~Corro.
\newblock Min{IE}: {M}inimizing {F}acts in {O}pen {I}nformation {E}xtraction.
\newblock In \emph{Proc. of the Conference on Empirical Methods in Natural
  Language Processing (EMNLP)}, pages 2630--2640, 2017.

\bibitem[Hoffart et~al.(2013)Hoffart, Suchanek, Berberich, and
  Weikum]{hoffart2013yago2}
Johannes Hoffart, Fabian~M Suchanek, Klaus Berberich, and Gerhard Weikum.
\newblock Y{AGO2}: {A} {S}patially and {T}emporally {E}nhanced {K}nowledge
  {B}ase from {W}ikipedia.
\newblock \emph{Artificial Intelligence}, 194:\penalty0 28--61, 2013.

\bibitem[Jain and Mausam(2016)]{jain2016knowledge}
Prachi Jain and Mausam.
\newblock Knowledge-{G}uided {L}inguistic {R}ewrites for {I}nference {R}ule
  {V}erification.
\newblock In \emph{Proc. of the Conference of the North American Chapter of the
  Association for Computational Linguistics: Human Language Technologies
  (NAACL-HLT)}, pages 86--92, 2016.

\bibitem[Lin et~al.(2012)Lin, Mausam, and Etzioni]{lin2012entity}
Thomas Lin, Mausam, and Oren Etzioni.
\newblock Entity {L}inking at {W}eb {S}cale.
\newblock In \emph{Proc. of the Joint Workshop on Automatic Knowledge Base
  Construction and Web-Scale Knowledge Extraction (AKBC-WEKEX@NAACL-HLT)},
  pages 84--88, 2012.

\bibitem[Manning et~al.(2014)Manning, Surdeanu, Bauer, Finkel, Bethard, and
  McClosky]{manning2014stanford}
Christopher Manning, Mihai Surdeanu, John Bauer, Jenny Finkel, Steven Bethard,
  and David McClosky.
\newblock The {S}tanford {C}ore{NLP} {N}atural {L}anguage {P}rocessing
  {T}oolkit.
\newblock In \emph{Proc. of the Annual Meeting of the Association for
  Computational Linguistics (ACL)}, pages 55--60, 2014.

\bibitem[Mausam(2016)]{mausam2016open}
Mausam.
\newblock Open {I}nformation {E}xtraction {S}ystems and {D}ownstream
  {A}pplications.
\newblock In \emph{Proc. of the International Joint Conference on Artificial
  Intelligence (IJCAI)}, pages 4074--4077, 2016.

\bibitem[Mausam et~al.(2012)Mausam, Schmitz, Bart, Soderland, and
  Etzioni]{schmitz2012open}
Mausam, Michael Schmitz, Robert Bart, Stephen Soderland, and Oren Etzioni.
\newblock Open {L}anguage {L}earning for {I}nformation {E}xtraction.
\newblock In \emph{Proc. of the Joint Conference on Empirical Methods in
  Natural Language Processing and Computational Natural Language Learning
  (EMNLP-CoNLL)}, pages 523--534, 2012.

\bibitem[Mitchell et~al.(2018)Mitchell, Cohen, Hruschka, Talukdar, Yang,
  Betteridge, Carlson, Dalvi, Gardner, Kisiel, Krishnamurthy, Lao, Mazaitis,
  Mohamed, Nakashole, Platanios, Ritter, Samadi, Settles, Wang, Wijaya, Gupta,
  Chen, Saparov, Greaves, and Welling]{mitchell2018never}
Tom Mitchell, William Cohen, Estevam Hruschka, Partha Talukdar, Bo~Yang, Justin
  Betteridge, Andrew Carlson, B.~Dalvi, Matt Gardner, Bryan Kisiel, Jayant
  Krishnamurthy, Ni~Lao, Kathryn Mazaitis, Thahir Mohamed, Ndapandula
  Nakashole, Emmanouil~A. Platanios, Alan Ritter, Mehdi Samadi, Burr Settles,
  Richard~C. Wang, Derry Wijaya, Abhinav Gupta, Xinlei Chen, Abulhair Saparov,
  Malcolm Greaves, and Joel Welling.
\newblock Never-{E}nding {L}earning.
\newblock \emph{Communications of the ACM}, 61\penalty0 (5):\penalty0 103--115,
  2018.

\bibitem[Moro and Navigli(2012)]{moro2012wisenet}
Andrea Moro and Roberto Navigli.
\newblock Wi{S}e{N}et: {B}uilding a {W}ikipedia-based {S}emantic {N}etwork with
  {O}ntologized {R}elations.
\newblock In \emph{Proc. of the ACM International Conference on Information and
  Knowledge Management (CIKM)}, pages 1672--1676, 2012.

\bibitem[Moro and Navigli(2013)]{moro2013integrating}
Andrea Moro and Roberto Navigli.
\newblock Integrating {S}yntactic and {S}emantic {A}nalysis into the {O}pen
  {I}nformation {E}xtraction {P}aradigm.
\newblock In \emph{Proc. of the International Joint Conferences on Artificial
  Intelligence (IJCAI)}, pages 2148--2154, 2013.

\bibitem[Nakashole et~al.(2012)Nakashole, Weikum, and
  Suchanek]{nakashole2012patty}
Ndapandula Nakashole, Gerhard Weikum, and Fabian Suchanek.
\newblock P{ATTY}: {A} {T}axonomy of {R}elational {P}atterns with {S}emantic
  {T}ypes.
\newblock In \emph{Proc. of the Joint Conference on Empirical Methods in
  Natural Language Processing and Computational Natural Language Learning
  (EMNLP-CoNLL)}, pages 1135--1145, 2012.

\bibitem[Riedel et~al.(2013)Riedel, Yao, McCallum, and
  Marlin]{riedel2013relation}
Sebastian Riedel, Limin Yao, Andrew McCallum, and Benjamin~M. Marlin.
\newblock Relation {E}xtraction with {M}atrix {F}actorization and {U}niversal
  {S}chemas.
\newblock In \emph{Proc. of the Conference of the North American Chapter of the
  Association for Computational Linguistics: Human Language Technologies
  (NAACL-HLT)}, pages 74--84, 2013.

\bibitem[Saur{\i} et~al.(2006)Saur{\i}, Littman, Knippen, Gaizauskas, Setzer,
  and Pustejovsky]{sauri2006timeml}
Roser Saur{\i}, Jessica Littman, Bob Knippen, Robert Gaizauskas, Andrea Setzer,
  and James Pustejovsky.
\newblock Time{ML} {A}nnotation {G}uidelines {V}ersion 1.2.1, 2006.

\bibitem[Shi and Weninger(2018)]{shi2018open}
Baoxu Shi and Tim Weninger.
\newblock Open-{W}orld {K}nowledge {G}raph {C}ompletion.
\newblock In \emph{Proc. of the Association for the Advancement of Artificial
  Intelligence (AAAI)}, pages 1957--1964, 2018.

\bibitem[Stanovsky and Dagan(2016)]{stanovsky2016creating}
Gabriel Stanovsky and Ido Dagan.
\newblock Creating a {L}arge {B}enchmark for {O}pen {I}nformation {E}xtraction.
\newblock In \emph{Proc. of the Conference on Empirical Methods in Natural
  Language Processing (EMNLP)}, pages 2300--2305, 2016.

\bibitem[Stanovsky et~al.(2015)Stanovsky, Dagan, and Mausam]{stanovsky2015open}
Gabriel Stanovsky, Ido Dagan, and Mausam.
\newblock Open {IE} as an {I}ntermediate {S}tructure for {S}emantic {T}asks.
\newblock In \emph{Proc. of the Annual Meeting of the Association for
  Computational Linguistics (ACL)}, volume~2, pages 303--308, 2015.

\bibitem[Vashishth et~al.(2018)Vashishth, Jain, and
  Talukdar]{vashishth2018cesi}
Shikhar Vashishth, Prince Jain, and Partha Talukdar.
\newblock C{ESI}: {C}anonicalizing {O}pen {K}nowledge {B}ases using
  {E}mbeddings and {S}ide {I}nformation.
\newblock In \emph{Proc. of the World Wide Web Conferences (WWW)}, pages
  1317--1327, 2018.

\bibitem[Wu et~al.(2018)Wu, Wu, Kao, and Yin]{wu2018towards}
Tien-Hsuan Wu, Zhiyong Wu, Ben Kao, and Pengcheng Yin.
\newblock Towards {P}ractical {O}pen {K}nowledge {B}ase {C}anonicalization.
\newblock In \emph{Proc. of the ACM Conference on Information and Knowledge
  Management (CIKM)}, pages 883--892, 2018.

\bibitem[Yahya et~al.(2014)Yahya, Whang, Gupta, and Halevy]{yahya2014renoun}
Mohamed Yahya, Steven Whang, Rahul Gupta, and Alon Halevy.
\newblock Re{N}oun: {F}act {E}xtraction for {N}ominal {A}ttributes.
\newblock In \emph{Proc. of the Conference on Empirical Methods in Natural
  Language Processing (EMNLP)}, pages 325--335, 2014.

\bibitem[Zaharia et~al.(2016)Zaharia, Xin, Wendell, Das, Armbrust, Dave, Meng,
  Rosen, Venkataraman, Franklin, Ghodsi, Gonzalez, Shenker, and
  Stoica]{zaharia2016apache}
Matei Zaharia, Reynold~S. Xin, Patrick Wendell, Tathagata Das, Michael
  Armbrust, Ankur Dave, Xiangrui Meng, Josh Rosen, Shivaram Venkataraman,
  Michael~J. Franklin, Ali Ghodsi, Joseph Gonzalez, Scott Shenker, and Ion
  Stoica.
\newblock Apache {S}park: {A} {U}nified {E}ngine {F}or {B}ig {D}ata
  {P}rocessing.
\newblock \emph{Communications of the ACM}, 59\penalty0 (11):\penalty0 56--65,
  2016.

\end{thebibliography}
\bibliographystyle{plainnat}

\clearpage
\appendix

\section{Further Alignments With DBpedia} \label{appendix:kb-hits}

\begin{table}[h]
	\footnotesize
	\centering
	\begin{tabular}{l@{\hspace{-.2cm}}rrr@{~}rcl@{\hspace{-.5cm}}r} \toprule
		Open         & Frequency in 	& Frequency in 	& \multicolumn2c{\# KB hits}& \# distinct  	& \multicolumn2c{Top-3 aligned DBpedia rel.}   	\\ 
		relation    & OPIEC-Clean  	& OPIEC-Link   	&           &              	& KB rels. 		& \multicolumn2c{and hit frequency} \\ \midrule
	      \textit{``leave''}& 130,515   	& 1,356     	&  347		& (25.6\%) 	& 54		& associatedBand & 70 \\ 
        			& 		&		& 		& 		& 		& associatedMusicalArtist& 70 \\ 
        			& 		& 		& 		& 		&   		& formerBandMember & 35 \\ \midrule
   \textit{``take''}		& 127,757	&  	660    	&   49      	& (7.4\%)	& 26 		& writer & 4 \\  
        			&		& 		&		& 		&		& previousWork & 4 \\ 
        			& 		& 		& 		& 		&		& artist & 3 \\ \midrule
    \textit{``use''}		& 127,537 	& 2,951 	& 123		& (4.2\%) 	& 53 		& currency & 9 \\ 
    				& 		& 		& 		& 		& 		& affiliation & 8 \\ 
    				& 		& 		& 		& 		& 		& timeZone & 7 \\ \midrule
\textit{``receive''}		& 118,429       & 2,133         & 268   	& (12.6\%) 	& 19   		& award & 236 \\
				& 		& 		& 		& 		& 	  	& team & 4 \\
				& 		& 		& 		& 		& 		& debutTeam & 4 \\ \midrule
\textit{``make''}   		& 116,688 	& 1,063       	& 140   	& (13.2\%)      & 39      	& director & 39 \\ 
				& 		& 		& 		& 		& 		& writer & 25 \\ 
				& 		&		& 		& 		& 		& producer & 12 \\ \midrule
\textit{``be member of''} 	& 104,680 	& 11,480        & 3,361  	& (29.3\%) 	& 80   		& associatedBand & 740 \\
				& 		& 		& 		& 		& 		& associatedMusicalArtist & 740 \\ 
				& 		& 		& 		& 		& 		& party & 584 \\ \midrule
\textit{``return to''}		&  104,392    	&    482        &  117      	&  (24.3\%)     &    35         & team & 44 \\
				&               &               &           	&               &               & league & 9 \\ 
				&               &               &           	&               &               & associatedBand & 7 \\ \midrule
\textit{``be at''}  		&  102,844      &  20,328       & 8,314     	&  (40.9\%)     &    78         & ground & 1,887 \\ 
				&               &               &           	&               &               & city & 1,759 \\ 
				&               &               &           	&               &               & location & 1,109 \\ \midrule
\textit{``be species of''}	& 101,846 	&   54,269      & 13,639    	&  (25.1\%)     &    9          & order & 5,196 \\ 
				&               &               &           	&               &               & family & 4,269 \\ 
				&               &               &           	&               &               & kingdom & 2,826 \\ \midrule
\textit{``move to''}		&  100,226      &    1,409      &  316      	&  (22.4\%)     &      43       & team & 124 \\ 
				&               &               &           	&               &               & managerClub & 35 \\ 
				&               &               &           	&               &               & ground & 16 \\ \midrule
\textit{``be write by''}	& 96,790    	&     6,340     &  1,956    	&   (30.9\%)    &      50       & author & 571 \\ 
				&               &               &           	&               &               & writer & 457 \\ 
				&               &               &           	&               &               & notableWork & 120 \\ \midrule 
\textit{``be found in''}	&  95,163   	&       836     &  110      	&    (13.2)\%   &      17       & location & 21 \\ 
				&               &               &           	&               &               & city & 19 \\ 
				&               &               &           	&               &               & headquarter & 17 \\ \bottomrule
 	\end{tabular}
	\caption{The most frequent open relations in OPIEC-Clean, along with DBpedia
          alignment information from OPIEC-Link (continuation of Tab.~\ref{tab:rel_ratios})}
        \label{tab:rel_ratios2}

\end{table}

\clearpage

\section{Complete List of OPIEC Meta-Data} \label{appendix:meta-data}


\begin{small}

\begin{longtable}{rp{11cm}}
	\centering
                                 &                                                                                \\[-.5cm] \toprule
		\textbf{Field}   & \textbf{Description}                                                           \\ \toprule
		Article ID 	 & Wikipedia article ID                                                           \\ \midrule
                Sentence         & Sentence from which the triple was extracted, including annotations:           
				  1) \textit{Sentence number} within the Wikipedia page;                         
                                  2) \textit{Span} of the sentence within the Wikipedia page;
				  2) \textit{Dependency parse};                                              
				  4) \textit{Token} information.                                                 
                                  For each token, OPIEC provides POS tag, NER type, span, the original word 
                                 found in the sentence, lemma, position of the                   
				 token within the sentence, and the WikiLink object (contains offset            
				  begin/end index of the link within the article, the original phrase of         
				  the link, and the link itself).                                                \\ \midrule
		Polarity 	 & The polarity of the triple (either \textit{positive} or \textit{negative})     \\ \midrule
		Negative words 	 & Words indicating negative polarity                                             \\ \midrule
		Modality 	 & The modality of the triple (either \textit{possibility} or \textit{certainty}) \\ \midrule
		CT/PS words 	 & Words indicating the detected modality                                         \\ \midrule
                Attribution      & Attribution of the triple (if found) including attribution phrase,             
                                  predicate, factuality, space and time                                          \\ \midrule
		Quantities 	 & Quantities in the triple (if found)                                            \\ \midrule
		Triple           & Lists of tokens with linguistic annotations for subject, predicate, and        
				  object of the triple                                                           \\ \midrule
		Dropped words    & To minimize the triple and make it more compact, MinIE sometimes               
                                  drops words considered to be semantically redundant words (e.g.,               
				  determiners). All dropped words  are stored here.                              \\ \midrule
		Time             & Temporal annotations, containing information about TIMEX3 type,                
				  TIMEX3 xml, disambiguated temporal expression, original core                   
				  words of the temporal expression, pre-modifiers/post-modifiers of              
				  the core words and temporal predicate                                          \\ \midrule
		Space            & Spatial annotations, containing information about the original                 
				  spatial words, the pre/post-modifiers and the spatial predicate                \\ \midrule
		Time/Space for phrases	 & Information about the temporal annotation on phrases. This                     
				  annotation contains:                                                           
				  1) \emph{modified word}: head word of the constituent being modified, and      
				  2) \emph{temporal/spatial} words modifying the phrase                          \\ \midrule
		Confidence score & The confidence score of the triple.                                            \\ \midrule
		Canonical links  & Canonical links for all links within the triple (follows redirections)         \\ \midrule
		Extraction type  & Either one of the clause types listed                                          
				  in ClausIE \cite{del2013clausie} (SVO, SVA, \ldots), or one of                 
				  the implicit extractions proposed in MinIE \cite{gashteovski2017minie}         
				  (Hearst patterns, noun phrases modifying persons, \ldots)  \\
		\bottomrule
\end{longtable}
\end{small}

\section{Spatio-Temporal Annotations} \label{appendix:SpaTe}

\subsection{Temporal Annotations on Triples}
\label{subsec:temporal-annotations-on-triples}

To provide temporal annotations, we use as raw input the input sentence, its
corresponding dependency parse tree, the $n$-ary extraction by
ClausIE~\cite{del2013clausie}, and the triple generated by
MinIE~\cite{gashteovski2017minie}. We run the SUTime temporal
tagger~\cite{chang2012sutime} to obtain a list of temporal expressions from the
input sentence. Next, we look at the head word $h$ of the triple's relation and
its descendants.

If $h$ has children in the dependency parse graph with typed dependency
\texttt{tmod} or \texttt{advmod}, then we check for each child $c$ if it is
contained in the list of temporal expressions. If so, we create a temporal
annotation for the triple with $c$ and all the descendants $c^+$ of $c$
in the dependency parse\footnote{We exclude descendants containing the
  typed dependencies \texttt{rcmod, punct, appos, dep, cc, conj} and
  \texttt{vmod}} (Figure
\ref{fig:TempAnnoExample1}).

\begin{figure}[h]
  \begin{adjustbox}{max width=\columnwidth}
  \parbox{14cm}{  
  		\hspace{4cm}
  		\begin{dependency}
		    \begin{deptext}
	     Bill \&  Gates \&  visited       \& Africa 	\& \textcolor{blue}{\textbf{last}} \& \textcolor{blue}{\textbf{week}}. \\
	       \&		\&	 	$h$		  \&   		\& 				$c^+$				   \&  				$c$		\\
		    \end{deptext}
		    \depedge{2}{1}{nn}
		    \depedge{3}{2}{nsubj}
		    \depedge{3}{4}{dobj}
		    \depedge{3}{6}{tmod}
		    \depedge{6}{5}{amod}
		    \deproot{3}{head word}
		  \end{dependency} \\ [.5ex]
  			 $~$ \hspace{0.8cm} $n$-ary extraction: (``Bill Gates'';  ``visited'';      ``Africa''  ``last week'') \\
  			 $~$ \hspace{0.95cm} Final extraction: (``Bill Gates'';  ``visited'';      ``Africa'')  \textcolor{blue}{\textbf{T: (last week)}}   \\
  }
  \end{adjustbox}
  \caption{Example of temporal annotation on triple with \texttt{tmod}}
  \label{fig:TempAnnoExample1}
\end{figure}

If $c$ is modified by the head word $h$ of the relation with a typed dependency
\texttt{prep}, then we check if the child of $c$ is a temporal expression. If
so, we note the temporal annotation of the triple with $c^+$ and its
descendants, and $c$ becomes the \emph{lexicalized temporal predicate} (Figure
\ref{fig:TempAnnoExample2}).

\begin{figure}[h]
\begin{adjustbox}{max width=\columnwidth}
  \parbox{14cm}{  
  		\hspace{4cm}
  		\begin{dependency}
		    \begin{deptext}
	     Bill \&  Gates \&  founded       \& Microsoft \& \textcolor{blue}{\textbf{in}} \& \textcolor{blue}{\textbf{\textbf{1975}}.} \\
	          \& 	    \& 		$h$ 	  \&    	   \&  		$c$						\&		$c^+$ \\
		    \end{deptext}
		    \depedge{2}{1}{nn}
		    \depedge{3}{2}{nsubj}
		    \depedge{3}{4}{dobj}
		    \deproot{3}{head word}
		    \depedge{3}{5}{prep}
		    \depedge{5}{6}{pobj}
		  \end{dependency} \\[.5ex]
  			$~$ \hspace{0.01cm} $n$-ary extraction: (``Bill Gates'';  ``founded''; ``Microsoft \textcolor{blue}{\textbf{in 1975}}'') \\
  			$~$ \hspace{0.2cm} Final extraction: (``Bill Gates'';  ``founded''; ``Microsoft'')  \textcolor{blue}{\textbf{T: (pred=in, t=1975)}}   \\
  }
  \end{adjustbox}
  \caption{Example of temporal annotation on triple with \texttt{prep}}
  \label{fig:TempAnnoExample2}
\end{figure}

If $c$ is modified by the head word $h$ with \texttt{xcomp}, then we check for temporal annotations the same
way as for \texttt{tmod} and \texttt{advmod}. However, we also treat $c^+$ as though is the relation head
word itself, so we also check if $c^+$'s children obey the same rules as for \texttt{tmod, advmod, prep} (Fig. \ref{fig:TempAnnoExample3}).

\begin{figure}
\begin{adjustbox}{max width=\columnwidth}
	  \parbox{14cm}{
	  		\hspace{4cm}  
	  		\begin{dependency}
			    \begin{deptext}
		     Elon \&  Musk \&  decided       \& to \& \underline{go} \& to \& Washington \& \textcolor{blue}{\textbf{yesterday}}. \\
		          \&       \&    $h$         \&    \&     $c$ 		 \&    \& 			 \&   				$c^+$ \\
			    \end{deptext}
			    \depedge{2}{1}{nn}
			    \depedge{3}{2}{nsubj}
			    \depedge{3}{5}{xcomp}
			    \depedge{5}{4}{aux}
			    \deproot{3}{head word}
			    \depedge{5}{6}{prep}
			    \depedge{6}{7}{pobj}
			    \depedge{5}{8}{tmod}
			  \end{dependency} \\ [.5ex]
	  			$n$-ary extraction: (``Elon Musk''; ``decided''; ``to go to Washington \textcolor{blue}{\textbf{yesterday}}'') \\
	  			Final extraction: (``Elon Musk''; ``decided to go to''; ``Washington'')  \textcolor{blue}{\textbf{T: (yesterday)}}   \\
	  }
  \end{adjustbox}
  \caption{Example of temporal annotation on triple with \texttt{xcomp}}
  \label{fig:TempAnnoExample3}
\end{figure}

Finally, MinIE-SpaTe tries to reason about how to construct the triples in a more compact manner. First,
we utilize ClausIE's $n$-ary extractions, and then see if $n > 3$. If so, we check if the whole temporal
extraction is one phrase, and if it is, then we drop it from the original triple, and proceed with the 
triples' structuring algorithm in MinIE (Figure \ref{fig:TempAnnoExample1}). 


\subsection{Temporal Annotations on Arguments} 
\label{subsec:temporal-annotations-on-args}
Sometimes the arguments themselves contain some temporal information
that is about the phrase itself, but not for the whole triple. Consider the sentence: \textit{``Isabella II 
opened the \textcolor{blue}{17th-century} Parque del Retiro \textit{in 1868}.''}. MinIE would extract the following triple:
	\textit{(``Isabella II''; ``opened the \textcolor{blue}{\textbf{17th-century}} Parque del Retiro
		\textcolor{blue}{\textbf{in}}''; ``\textcolor{blue}{\textbf{1868}}'')}.
This triple contains two pieces of temporal information: \textit{``1868''} and \textit{``17th-century''}. MinIE-SpaTe rewrites this as:
	 \textit{(``Isabella II''; ``opened''; ``Parque del Retiro'') 
	 	\textcolor{blue}{T: (in, 1868);} Object $\rightarrow$ \textcolor{blue}{T: 17th-century}}

Distinguishing these types of temporal information is important, because the correct temporal
information should be attached to the correct place. In this example, we should not attach \textit{``17th century''}
as a temporal annotation to the whole triple (because the park was not opened in 17th century by Isabella II), but just to the object
(the park \emph{itself} is from 17th century).

To obtain such annotations, we search for a noun or an adjective in a phrase
(word $w$), and check if it is modified by one the following dependencies:
\texttt{amod}, \texttt{acmop}, \texttt{advmod}, \texttt{nn}, \texttt{num},
\texttt{number}, \texttt{tmod}. If so, we consider the children of $w$. If a
child of $w$ is a temporal word, we include it in the temporal annotation and
drop it from the triple.\footnote{We ignore the children derived from the
  following dependencies: \texttt{rcmod}, \texttt{punct}, \texttt{appos},
  \texttt{cc}, \texttt{conj}}

\subsection{Spatial Annotations}

For spatial annotations, we follow a similar approach as for temporal
annotations. The main differences are (1) instead of getting a list of annotated
temporal expressions, we get a list of locations from Stanford
NER~\cite{finkel2005incorporating}, and (2) we use the similar rules as for
temporal annotations for \texttt{prep}, but not for \texttt{xcomp},
\texttt{tmod}, and \texttt{advmod} dependencies.

\subsection{Precision of Spatio-Temporal Annotations}

We performed an experimental study to assess the precision of the
spatio-temporal annotations. We used the sample of 10k random sentences drawn
from the New York Times corpus by~\citet{gashteovski2017minie}. We ran
MinIE-SpaTe on these sentences and created four sets of extractions: triples
containing temporal annotations on (1a) the triple level and (1b) the argument
level, and triples containing spatial annotations on (2a) the triple level and
(2b) the argument level. From each subset, we selected 200 random triples and
labeled whether the corresponding MinIE extraction (which does not provide
spatio-temporal annotations) was correctly extracted. We construct our final
evaluations sets by including 100 random but correctly extracted triples. In
each subset, a human labeler then assessed whether the spatial and/or temporal
annotations provided by MinIE-SpaTe were correct as well.

For the triple-level temporal annotations, 91/100 extractions were labeled as
correctly annotated; for the argument-level temporal annotations, 80/100 were
correctly annotated. Similar precision was measured for the spatial annotations:
91/100 on the triple-level and 82/100 on the argument level. We performed an
error analysis and found that common errors stem from incorrect spatio-temporal
tags (e.g., the argument \textit{``Summer Olympics''} gets reduced to
\textit{``Olympics''} with temporal annotation \textit{``Summer''}) or incorrect
dependency parses.


\section{Confidence Score Features} \label{appendix:conf}


\begin{table}[h]
	\footnotesize
	\centering
	\begin{tabular}{rr} \toprule
		Feature 								& 	Value 		\\ \midrule
		Length (sentence / extraction)			& 	int / int	\\ 
		Clause type 							& SVA, SVO, ...	\\ 
		Dropped all optional adverbials / prep.	& bool / bool	\\ 
		Relation as whole string in sentence	& bool			\\ 
		Comparison of POS Tags					& bool 			\\  
		of conjunction words in subject 		& 				\\ 
		Contains possessive relation / gerund	& bool / bool	\\ 
		Contains infinitive verb in subj / rel	& bool / bool	\\ 
		Order of words in the extraction 		& bool 			\\ 
		Extraction contains \texttt{dep}        & bool 			\\ 
		Processed conjunction subj/rel/obj 		& bool / bool / bool \\ 
		Object before subject in sentence 		& bool 			\\ 
		Triple occurs in MinIE-D / MinIE-A		& bool / bool	\\ 
		Is the relation frequent?\footnotemark	& bool 			\\  		
		Extracts quantity / time / space		& bool / bool / bool \\ \bottomrule
 	\end{tabular}
\end{table}
\footnotetext{minimum support: 100 K}







\end{document}